\documentclass{sip}
\usepackage{epstopdf, epsfig} 
\usepackage{amsmath}
\usepackage{amsfonts}
\usepackage{array}
\usepackage{graphicx}
\usepackage{booktabs}
\usepackage{hyperref}
\usepackage{natbib}

\title{Graph Representation Learning: A Survey}

\author{Fenxiao Chen, Yuncheng Wang, Bin Wang and C.-C. Jay Kuo}

\address{University of Southern California, Los Angeles, CA 90089, USA}

\corres{\name{Fenxiao Chen}
\email{fenxiaoc@usc.edu}}

\begin{document}

\begin{abstract}
Research on graph representation learning has received a lot of
attention in recent years since many data in real-world applications
come in form of graphs. High-dimensional graph data are often in
irregular form, which makes them more difficult to analyze than
image/video/audio data defined on regular lattices. Various graph
embedding techniques have been developed to convert the raw graph data
into a low-dimensional vector representation while preserving the
intrinsic graph properties. In this review, we first explain the graph
embedding task and its challenges.  Next, we review a wide range of
graph embedding techniques with insights. Then, we evaluate several
stat-of-the-art methods against small and large datasets and compare
their performance.  Finally, potential applications and future
directions are presented. 
\end{abstract}


\maketitle

\section{Introduction}\label{sec:introduction}

Research on graph representation learning has gained more and more
attention in recent years since many real world data can be represented
by graphs conveniently.  Examples include social networks, linguistic
(word co-occurrence) networks, biological \cite{theocharidis2009network}
networks and many other multimedia domain-specific data. Graph
representation allows the relational knowledge of interacting entities
to be stored and accessed efficiently \cite{angles2008survey}. Analysis
of graph data can provide significant insights into community detection
\cite{gargi2011large}, behavior analysis and other useful applications
such as node classification \cite{bhagat2011node}, link prediction
\cite{liben2007link} and clustering \cite{goyal2018graph}.  Various
graph embedding techniques have been developed to convert the raw graph
data into a high-dimensional vector while preserving intrinsic graph
properties. This process is also known as graph representation learning.
With a learned graph representation, one can adopt machine learning
tools to perform downstream tasks conveniently. 
Obtaining an accurate representation of a graph is challenging in three
aspects. First, finding the optimal embedding dimension of a
representation \cite{yan2005graph}, \cite{yan2007graph} is not an easy
task. A representation of a higher dimension tends to preserve more
information of the original graph at the cost of storage and
computation.  A representation of a lower dimension is more resource
efficient.  It may reduce noise in the original graph as well.  However,
there is a risk in losing some critical information of the original
graph.  The dimension choice depends on the input graph type as well as
the application domain \cite{shaw2009structure}. Second, choosing the
proper graph property to embed is an issue of concern if a graph has a
plethora of properties. Third, many graph embedding methods have been
developed in the past.  It is desired to have some guideline in
selecting a good embedding method for a target application. 
In this work, we intend to provide an extensive survey on graph
embedding methods with the following three contributions in mind. 
\begin{itemize}
\item We would like to offer new comers in this field a global
perspective with insightful discussion and an extensive reference list.
Thus, a wide range of graph embedding techniques, including the most
recent graph representation models, are reviewed. 
\item To shed light on the performance of different embedding methods,
we conduct extensive performance evaluation on both small and large data
sets in various application domains. To the best of our knowledge, this
is the first survey paper that provides systematic evaluation of a rich
set of graph embedding methods in domain-specific applications. 
\item We provide an open-source Python library, called the Graph
Representation Learning Library (GRLL), to readers. It offers a unified
interface for all graph embedding methods discussed in this paper.  This
library covers the largest number of graph embedding techniques up to
now.
\end{itemize}
The rest of this paper is organized as follows. We first state the
problem as well as several definitions in Sec.  \ref{sec:definition}.
Then, traditional and emerging graph embedding methods are reviewed in
Sec.  \ref{sec:techniques} and \ref{sec:e-techniques}, respectively.
Next, we conduct extensive performance evaluation on a large number of
embedding methods against different datasets in different application
domains in Sec. \ref{sec:evaluation}.  The application of the learned
graph representation and the future research directions are discussed in
Sec. \ref{sec:applications} and Sec.  \ref{sec:future}, respectively.
Finally, concluding remarks are given in Sec. \ref{sec:conclusion}. 

\section{Definition and Preliminaries}\label{sec:definition}

\subsection{Notations}

A graph, denoted by $G=(V,E)$, consists of vertices, $V =
\{v_1,v_2,...,v_n\}$, and edges, $E = \{e_{i,j}\}$, where an edge
$e_{i,j}$ connects vertex $v_i$ to vertex $v_j$. Graphs are usually
represented by an adjacency matrix or a derived vector space
representation \cite{ding2001min}.  The adjacency matrix, $A$, of graph
$G$ contains non-negative weights associated with each edge, $a_{ij}
\geq 0 $. If $v_i$ and $v_j$ are not directly connected to one another,
$a_{ij} = 0$.  For undirected graphs, $a_{ij} = a_{ji} $ for all $1 \leq
i \leq j \leq n$. 

Graph representation learning (or graph embedding) aims to map each node
to a vector where the distance characteristics among nodes is preserved.
Mathematically, for graph $G=(V,E)$, we would like to find a mapping:
$$
f:v_i \rightarrow x_i\in \mathbb{R}^{d},
$$
where $d \ll |V|$, and $X_i = \{x_1,x_2,...,x_d\}$ is the embedded (or 
learned) vector that captures the structural properties of vertex $v_i$. 

The first-order proximity \cite{cavallari2017learning} in a network is
the pairwise proximity between vertices. For example, in weighted
networks, the weights of the edges are the first-order proximity between
vertices. If there is no edge observed between two vertices, the
first-order proximity between them is 0. If two vertices are linked by
an edge with high weight, they should be close in the embedding space.
This objective can be obtained by minimizing the distance between the
joint probability distribution in the vector space and the empirical
probability distribution of the graph.  If we use the KL-divergence
\cite{goldberger2003efficient} to calculate the distance, the objective
function is given by
\begin{equation}\label{eq:1}
O_1 = -\sum_{(i,j) \in E}w_{ij}\log p_1(v_i,v_j),
\end{equation}
where 
$$
p_1(v_i,v_j) = \frac{1}{1+\exp(\vec{-u_i}^T \cdot \vec{u_j})}
$$ 
and where $\vec{u_i} \in \mathbb{R}^{d}$ is the low-dimensional vector
representation of vertex $v_i$ and $w_{ij}$ is the weight.  The
second-order proximity \cite{zhou2017scalable} is used to capture the
2-step relationship between two vertices. Although there is no direct
edge between two vertices of the second-order proximity, their
representation vectors should be close in the embedded space if they
share similar neighborhood structures. 

The objective function of the second-order proximity can be defined as
\begin{equation}\label{eq:2}
O_2 = -\sum_{(i,j) \in E}w_{ij}\log p_2(v_i|v_j),
\end{equation}
where $w_{i,j}$ is the edge weight between node $i$ and $j$ and
\begin{equation}
p_2(v_j|v_i = \frac{\exp(\vec{u_j^{\,'}}^T \cdot \vec{u_i})}{\sum_{k = 1}^{|V|} 
\exp(\vec{u_k^{\,'}}^T \cdot \vec{u_i})},
\end{equation}
and where $\vec{u_i^{\,}} \in \mathbb{R}^{d}$ is the representation of
vertex $v_i$ when it is treated as a vertex and $\vec{u_j^{\,'}} \in
\mathbb{R}^{d}$ is the vector representation of vertex $v_j$ when it is
treated as a specific context for vertex $v_i$. 

Graph sampling is used to simplify graphs \cite{anis2016efficient}. They can be 
categorized into two types.
\begin{itemize}
\item Negative Sampling \cite{mikolov2013distributed},\cite{xu2015semantic}  \\
Negative sampling is proposed as an alternative to the hierarchical
computation of the softmax, which reduces the runtime of the softmax
computation on a large scale dataset. Graph optimization requires the
summation over the entire set of vertices.  It is computationally
expensive for large-scale networks.  Negative sampling is developed to
address this problem. It helps distinguish the neighbors from other
nodes by sampling multiple negative samples according to the noise
distribution. In the training process, correct surrounding neighbors
positive examples in contrast to a set of sampled negative examples
(usually noise). 
\item Edge Sampling \cite{leskovec2006sampling}, \cite{ribeiro2010estimating} \\ 
In the training stage, it is difficult to choose an appropriate learning
rate in graph optimization when the difference between edge weights is
large. To address this problem, one solution is to use edge sampling
that unfolds weighted edges into several binary edges at the cost of
increased memory.  An alternative is treating weighted edges as binary
ones with their sampling probabilities proportional to the weights.
This treatment would not modify the objective function. 
\end{itemize}

\subsection{Graph Input} 

Graph embedding methods take a graph as the input, where the graph can
be homogeneous graphs, heterogeneous graphs, with/without auxiliary
information or constructed graphs \cite{cai2018comprehensive}. They are
detailed below. 
\begin{itemize}
\item Homogeneous graphs refer to graphs whose nodes and edges belong to
the same type. All nodes and edges of homogeneous graphs are treated
equally. 
\item Heterogeneous graphs contain different edge types to represent
different relations among different entities or categories. For example,
their edges can be directed or undirected.  Heterogeneous graphs
typically exist in community-based question answering (cQA) sites,
multimedia networks and knowledge graphs.  Most social network graphs
are directed graphs \cite{tang2015line}. Only the basic structural 
information of input graphs is provided in real world applications.
\item Graphs with auxiliary information \cite{gilbert2004compressing},
\cite{taylor2001heterogeneous} are those that have labels, attributes,
node features, information propagation, etc. A label indicates node's
category. Nodes with different labels should be embsedded further away
than those with the same label. An attribute is a discrete or continuous
value that contains additional information about the graph rather than
just the structural information.  Node features are shown in form of
text information for each node.  Information propagation indicates
dynamic interaction among nodes such as post sharing or "retweet" while
Wikipedia \cite{yang2015multi}, DBpedia \cite{bizer2009dbpedia},
Freebase \cite{bollacker2008freebase}, etc. provide popular knowledge
bases. 
\item Graphs constructed from non-relational data are assumed to lie in a
low dimensional manifold. 
\end{itemize}

The input feature matrix can be represented as $X \in \mathbb{R}^ {{|V|}
\times {N}} $ where each row $X_i$ is a $N$-dimensional feature vector
for the $i^{th}$ training instance. A similarity matrix, denoted by $S$,
can be constructed by computing the similarity between $X_i$ and $X_j$
for graph classifications.

\subsection{Graph Output}

The output of a graph embedding method is a set of vectors representing
the input graph. It could be node embedding, edge embedding, hybrid
embedding or whole-graph embedding. The preferred output form is
application-oriented and task-driven. We elaborate them below.
\begin{itemize}
\item Node embedding represents each node as a vector, which would be useful
for node clustering and classification. For node embedding, nodes that
are close in the graph are embedded closer together in the vector
representations.  Closeness can be first-order proximity, second-order
proximity or other similarity calculation.  
\item Edge embedding aims to map each edge into a vector.  It is useful
for predicting whether a link exists between two nodes in a graph.  For
example, knowledge graph embedding can be used for knowledge graph
entity/relation prediction. 
\item Hybrid embedding is the combination of different types of graph
components such as node and edge embedding. Hybrid embedding is useful
for semantic proximity search and subgraphs learning. It can also be
used for graph classification based on graph kernels. Substructure or
community embedding can also be done by aggregating individual node and
edge embedding inside it. Sometimes, better node embedding is learned
by incorporating hybrid embedding methods. 
\item Whole graph embedding is usually done for small graphs such as
proteins and molecules. These smaller graphs are represented as one
vector, and two similar graphs are embedded to be closer.  Whole-graph
embedding facilitates graph classification tasks by providing a
straightforward and efficient solution in computing graph similarities. 
\end{itemize}

\subsection{History of Graph Embedding}

The study of graph embedding can be traced back to 1900s when people
questioned whether all planar graphs with $n$ vertices have a straight
line embedding in an $n_k\times n_k$ grid. This problem was solved in
\cite{de1988small} and \cite{fary1948straight}.  The same result for
convex maps was proved in \cite{stein1951convex}.  More analytic work on
the embedding method and time/space complexity of such a method were
studied in \cite{chrobak1995linear} and \cite{de1990draw}.  However, a
more general approach is needed since most real world graphs are not
planer. A large number of methods have been proposed since then. 

\subsection{Overview of Graph Embedding Ideas}

We provide an overview on various graph embedding ideas below.
\begin{itemize}
\item Dimensionality Reduction \\
In early 2000s, graph embedding is achieved by dimensionality reduction.
For a graph with $n$ nodes, each of which is of dimension $D$, these
embedding methods aim to embed nodes into a $d$-dimensional vector
space, where $d \ll D$. They are called classical methods and reviewed
in Section \ref{CM}. Dimensionality reduction is less scalable. 
\item Random Walk \\
One can trace a graph by starting random walks from random initial nodes
so as to create multiple paths. These paths reveal the context of
connected vertices. The randomness of these walks gives the ability to
explore the graph, capture the global and local structural information
by walking through neighboring vertices.  Later on, probability models
like skip-gram and bag-of-word are performed on the random sampled paths
to learn node representations.  The random walk based methods will
be discussed in Section \ref{RWBM}. 
\item Matrix Factorization \\
By leverage the sparsity of real-world networks, one can apply the
matrix factorization technique that finds an approximation matrix for
the original graph.  This idea is elaborated in Section \ref{MFBM}. 
\item Neural Networks \\
Neural network models such as convolution neural network (CNN)
\cite{krizhevsky2012imagenet}, recursive neural networks (RNN)
\cite{mikolov2010recurrent} and their variants have been widely adopted
in graph embedding. 
This topic will be described in Section \ref{GCN}. 
\item Large Graphs \\
Some large graphs are difficult to embed since CNN and RNN 
models do not scale well with the numbers of edges and
nodes. New embedding methods are designed targeting at large graphs.
They become popular due to their efficiency. This topic is reviewed in
Section \ref{LGEM}
\item Hypergraphs \\
Most social networks are hypergraphs. As social networks get more
attention in recent years, hypergraph embedding becomes a hot
topic, which will be presented in Section \ref{HyperE}. 
\item Attention Mechanism \\
Attention mechanism can be added to existing embedding models to
increase embedding accuracy, which will be examined in Section \ref{LGEM}. 
\end{itemize}
An extensive survey on graph embedding methods will be conducted in the
next section. 

\section{Classical Methods}\label{sec:techniques}

\subsection{Dimension-Reduction-Based Methods}\label{CM}

Classical graph embedding methods aim to reduce the dimension of
high-dimensional graph data into a lower dimensional representation
while preserving the desired properties of the original data.  They can
be categorized into linear and nonlinear two types.  The linear methods
include the following. 
\begin{itemize}
\item Principal Component Analysis (PCA) \cite{jolliffe2011principal} \\
The basic assumption for PCA is that that principal components that
are associated with larger variances represent the important structure
information while those smaller variances represent noise.  Thus, PCA
computes the low-dimensional representation that maximizes the data
variance. Mathematically, it first finds a linear transformation matrix
$W \in \mathbb{R}^{D\times d}$ by solving
\begin{equation}
W =\mbox{argmax} \mbox{Tr} (W^T \mbox{Cov} (X)W), \quad d=1, 2, \cdots, D,
\end{equation}
where $\mbox{Cov} (X)$ denotes the covariance of data matrix $X$.  
It is well known that the principal components are orthogonal and they
can be solved by eigen decomposition of the covariance of data matrix
\cite{umeyama1988eigendecomposition}. 
\item Linear Discriminant Analysis (LDA) \cite{ye2005two} \\
The basic assumption for LDA is that each class is Gaussian distributed.
Then, the linear projection matrix, $W \in \mathbb{R}^{D\times d}$, can
be obtained by maximizing the ratio between the inter-class scatter and
intra-class scatters. The maximization problem can be solved by eigen
decomposition and the number of low dimension $d$ can be obtained by
detecting a prominent gap in the eigen-value spectrum. 
\item Multidimensional Scaling (MDS) \cite{robinson1995typology} \\ 
MDS is a distance-preserving manifold learning method. It preserves
spatial distances. MDS derives a dissimilarity matrix $D$, where $D^{i,j}$
represents the dissimilarity between points $i$ and $j$, and produces a
mapping in a lower dimension to preserve dissimilarities as much as
possible. 
\end{itemize}

The above three-mentioned methods are referred to as ``subspace
learning" \cite{yan2007graph} under the linear assumption. However,
linear methods might fail if the underlying data are highly non-linear
\cite{saul2006spectral}. Then, non-linear dimensionality reduction
(NLDR) \cite{demers1993non} can be used for manifold learning. The objective
is to learn the nonlinear topology automatically. The NLDR methods include
the following.
\begin{itemize}
\item Isometric Feature Mapping (Isomap) \cite{samko2006selection} \\
The Isomap finds low-dimensional representation that most accurately
preserves the pairwise geodesic distances between feature vectors in all
scales as measured along the submanifold from which they were sampled.
Isomap first constructs neighborhood graph on the manifold, then it
computes the shortest path between pairwise points.  Finally it
constructs low-dimensional embedding by applying MDS. 
\item Locally Linear Embedding (LLE) \cite{roweis2000nonlinear} \\
LLE preserves the local linear structure of nearby feature vectors. LLE
first assign neighbors to each data point. Then it compute the weights
$W^{i,j}$ that best linearly reconstruct Xi from its neighbors. Finally
it compute the low-dimensional embedding that best reconstructed by
$W^{i,j}$. Besides NLDR, kernel PCA is another dimension reduction
technique that is comparable to Isomap, LLE. 
\item Kernel Methods \cite{harandi2011graph} \\
Kernel extension can be applied to algorithms that only need to compute
the inner product of data pairs. After replacing the inner product with
kernel function, data is mapped implicitly from the original input space
to higher dimensional space and then apply linear algorithm in the new
feature space. The benefit of kernel trick is data that are not linearly
separable in the original space could be separable in new high
dimensional space. Kernel PCA is often used for NLDR with polynomial or
Gaussian kernels. 
\end{itemize}

\subsection{Random-Walk-Based Methods}\label{RWBM}

Random-walk-based methods sample a graph with a large number of paths by
starting walks from random initial nodes. These paths indicate the
context of connected vertices. The randomness of walks gives the ability
to explore the graph and capture global as well as local structural
information by walking through neighboring vertices.  After that,
probability models such as skip-gram and bag-of-word can be performed on
these randomly sampled paths to learn the node representation. 

\begin{itemize}
\item DeepWalk \cite{perozzi2014deepwalk} \\
DeepWalk is the most popular random-walk-based graph embedding method.
In DeepWalk, a target vertex, $v_i$, is said to belong to a sequence $S
= \{ v_1, \cdots ,v_{|s|}\}$ sampled by random walks if $v_i$ can reach
any vertex in $S$ within a certain number of steps. The set of vertices,
$V_s = \{ v_{i-t}, \cdots ,v_{i-1},v_{i+1}, \cdots, v_{i+t}\}$, is the
context of center vertex $v_i$ with a window size of $t$. DeepWalk aims
to maximize the average logarithmic probability of all vertex context
pairs in random walk sequence $S$. It can be written as
\begin{equation}
\frac{1}{|S|}\sum_{i=1}^{|S|} \sum_{-t\leq j \leq t,j\neq 0} \log p(v_{i+j}|v_i),
\end{equation}
where $p(v_j|v_i)$ is calculated using the softmax function. It is
proven in \cite{yang2015network} that DeepWalk is equivalent to
factoring a matrix
\begin{equation}\label{eq:factorization}
M = W^T \times H,
\end{equation}
where $M\in \mathbb{R}^{|V|\times|V|}$ whose entry, $M_{ij}$, is the
logarithm of the average probability that vertex $v_i$ can reach vertex
$v_j$ in a fixed number of steps and $W\in \mathbb{R}^{k\times|V|}$ is
the vertex representation. The information in $H \in
\mathbb{R}^{k\times|V|}$ is rarely utilized in the classical DeepWalk
model. 
\item node2vec \cite{grover2016node2vec} \\
node2vec is a modified version of DeepWalk. In DeepWalk, sampled
sequences are based on DFS (Depth-first Sampling) strategy. They consist
of neighboring nodes sampled at increasing distances from the source
node sequentially. However, if the contextual sequences are sampled by
the DFS strategy alone, only few vertices close to the source node will
be sampled. Consequently, the local structure will be easily overlooked.
In contrast with the DFS strategy, the BFS (Breadth-first Sampling)
strategy will explore neighboring nodes with a restricted maximum
distance to the source node while the global structure may be neglected.
\\ As a result, node2vec proposes a probability model in which the
random walk has a certain probability, $\frac{1}{p}$, to revisit nodes
being traveled before. Furthermore, it uses an in-out parameter $q$ to
control the ability of exploring the global structure. When return
parameter $p$ is small, the random walk may get stuck in a loop and
capture the local structure only.  When in-out parameter $q$ is small,
the random walk is more similar to a DFS strategy and capable of
preserving the global structure in the embedding space. 
\end{itemize}


\subsection{Matrix-Factorization-Based Methods}\label{MFBM}

To obtain node embedding, matrix-factorization-based embedding methods,
also called Graph Factorization (GF) \cite{ahmed2013distributed}, was
the first one to achieve graph embedding in $O(|E|)$ time. To obtain the
embedding, GF factorizes the adjacency matrix of a graph. It corresponds
to a structure-preserving dimensionality reduction process.  There are
several variations as summarized below. 
\begin{itemize}
\item Graph Laplacian Eigenmaps \cite{belkin2003laplacian} \\
This technique minimizes a cost function to ensure that points close to
each other on the manifold are mapped close to each other in the
low-dimensional space to preserve local distances. 
\item Node Proximity Matrix Factorization \cite{singh2008relational} \\
This method approximates node proximity in a low-dimensional space
via matrix factorization by minimizing the following objective function:
\begin{equation}\label{NPMF}
Y =\mbox{argmax} \min| W-YY^{T}|,
\end{equation}
where $W$ is the node proximity matrix, which can be derived
by several methods. One way to obtain $W$ is to use Eq. 
(\ref{eq:factorization}).
\item Text-Associated DeepWalk (TADW) \cite{yang2015network} \\
TADW is an improved DeepWalk method for text data. It incorporates the
text features of vertices in network representation learning via matrix
factorization. Recall that the entry, $m_{ij}$, of matrix $M \in
\mathbb{R}^{|V| \times |V|}$ denotes the logarithm of the average
probability that vertex $v_i$ randomly walks to vertex $v_j$. Then, TADW
factorizes $M$ into three matrices
\begin{equation}\label{factorization}
M = W^T \times H \times T,
\end{equation}
where $W\in \mathbb{R}^{k\times |V|}$, $H\in \mathbb{R}^{k\times f_t}$
and $T \in \mathbb{R}^{f_t\times |V|}$ is the text feature matrix. In
TADW, $W$ and $HT$ are concatenated as the representation for vertices.
\item Homophily, Structure, and Content Augmented (HSCA) Network
\cite{zhang2016homophily} \\
The HSCA model is an improvement upon the TADW model. It uses
Skip-Gram and hierarchical Softmax to learn a distributed word
representation. The objective function for HSCA can be written as
\begin{equation}\label{eq:HSCA}
\begin{split}
\underset{W,H}\min(||M-W^T H T||_F^2+\frac{\lambda}{2} (||W||_F^2 +||H||_F^2) \\
+ \mu(R_1(W)+R_2(H))),
\end{split}
\end{equation}
where $||.||_2$ is the matrix $l_2$ norm and $||.||_F$ is the matrix
Frobenius form. In Eq.  (\ref{eq:HSCA}), the first term aims to minimize
the matrix factorization error of TADW.  The second term imposes the
low-rank constraint on $W$ and $H$ and uses ${\lambda}$ to control the
trade-off. The last regularization term enforces the structural
homophily between connected nodes in the network.  The conjugate
gradient (CG) \cite{moller1993scaled} optimization technique can be used
to update $W$ and $H$. We may consider another regularization term 
to replace the third term; namely,
\begin{equation}
R(W,H) = \frac{1}{4} \sum_{i=1,j = 1}^{|V|} A_{i,j}||\begin{bmatrix} w_i \\
Ht_i \end{bmatrix}
-
\begin{bmatrix}
w_j \\
Ht_j
\end{bmatrix}||_2^2.
\end{equation}
This term will make connected nodes close to each other in the learned network 
representation \cite{chen2019deepwalk}.  
\item GraRep \cite{cao2015grarep} \\
GraRep aims to preserve the high order proximity of graphs in the
embedding space. While the random-walk based methods have a similar
objective, their probability model and objective functions used are difficult
to explain how the high order proximity are preserved. GraRep derives a
$k$-th order transition matrix, $A^k$, by multiply the adjacency matrix
to itself $k$ times. The transition probability from vertex $w$ to
vertex $c$ is the entry in the $w$-th row and $c$-th column of the
$k$-th order transition matrix. Mathematically, it can be written as
\begin{equation} \label{eq:grarep}
p_k(c|w) = A^k_{w,c}.
\end{equation}
With the transition probability defined in (\ref{eq:grarep}), the loss
function is defined by the skip-gram model and negative sampling. To
minimize the loss function, the embedding matrix can be expressed as
\begin{equation} \label{eq:rep}
Y^k_{i,j} = W_i^k \cdot C_j^k = \log(\frac{A^k_{i,j}}{\sum_{t}A^k_{t,j}}) - 
\log(\beta),
\end{equation}
where $\beta$ is a constant $\frac{\lambda}{N}$, $\lambda$ is the
negative sampling parameter, and $N$ is the number of vertices. The
embedding matrix, $W$, can be obtained by factorizing matrix $Y$ in
(\ref{eq:rep}). 
\item HOPE \cite{ou2016asymmetric} \\
HOPE preserves asymmetric transitivity in approximating the high order
proximity, where asymmetric transitivity indicates a certain correlation
among directed graphs. Generally speaking, if there is a directed edge
from $u$ to $v$, it is likely that there is a directed edge from $v$ to
$u$ as well.  Several high order proximities such as the Katz Index
\cite{katz1953new}, the Rooted PageRank, the Common Neighbors, and the
Adamic-Adar were experimented in \cite{ou2016asymmetric}. The embedding,
$\vec{v_i}$, for node $i$ can be obtained by factorizing the proximity
matrix, $S$, derived from these proximities. To factorize $S$, SVD is
adopted and only the top-k eigenvalues are chosen. 
\end{itemize}

\section{Emerging Methods}\label{sec:e-techniques}

\subsection{Neural-Network-Based Methods}\label{GCN}

Neural network models become popular again since 2010.  Being inspired
by the success of recurrent neural networks (RNNs) and convolutional
neural networks (CNNs), researchers attempt to generalize and apply them
to graphs.  Natural language processing (NLP) models often use the RNN
to find a vector representation for words.  The Word2Vec
\cite{mikolov2013efficient} and the skip-gram models
\cite{mikolov2013distributed} aim to learn the continuous feature
representation of words by optimizing a neighborhood preserving
likelihood function. By following this idea, one can adopt a similar
approach for graph embedding, leading to the Node2Vec method
\cite{grover2016node2vec}. Node2Vec utilizes random walks
\cite{spitzer2013principles} with a bias to sample the neighborhood of a
target node and optimizes its representation using stochastic gradient
descent (SGD). Another family of neural-network-based embedding methods
adopt CNN models. The input is either paths sampled from a graph or the
whole graph itself.  Some use the original CNN model designed for the
Euclidean domain and reformat the input graph to fit it. Others
generalize the deep neural model to non-Euclidean graphs. 

Several neural-network-based Methods based graph embedding methods are
presented below. 
\begin{itemize}
\item Graph Convolutional Network (GCN) \cite{kipf2016semi} \\
GCN allows end-to-end learning of the graph with arbitrary size and
shape.  This model uses convolution operator on the graph and
iteratively aggregates the embedding of neighbors for nodes. This
approach is widely used for semi-supervised learning on graph-structured
data that is based on an efficient variant of convolutional neural
networks that operates directly on graphs and it learns hidden layer
representations that encode both local graph structure and features of
nodes. In the first step of the GCN, a node sequence will be selected.
The neighborhood nodes will be assembled, then the neighborhood might be
normalized to impose order of the graph, then convolutional layers will
be used to learn representation of nodes and edges. The propagation rule
used is
\begin{equation}\label{eq:gcn1}
f(H^{(l)},A) = \sigma ({D}^{-\frac{1}{2}}{A}{D}^{-\frac{1}{2}}H^{(l)}W^{(l)}),
\end{equation}
where $A$ is the identity matrix, with enforced self loops to include
the node features of itself we get \^{$A$}$ = A + I$, $I$ is the
identity matrix and \^{$D$} is the diagonal node degree matrix of
\^{$A$} to normalize the neighbors of $A$.  Under spectral graph theory
of CNNs on graphs, GCN is equivalent to Graph Laplacian in the
non-euclidean domain \cite{defferrard2016convolutional}. The
decomposition of eigenvalues for the Normalized Graph Laplacian data can
also be used for tasks such as classification and clustering. However,
GCN usually only uses two convolutional layer and why it works its not
well explained. One recent work showed that GCN model is a special from
of Laplacian smoothing \cite{li2018deeper} \cite{field1988laplacian}.
This is the reason that GCN works, and also more than two convolutional
layers will lead to over-smoothing, therefore making making the features
of nodes similar to each other and more difficult to separate from each
other. 
\item Signed Graph Convolutional Network (SGCN) \cite{derr2018signed} \\
Most GCNs operate on unsigned graphs, however, many links in real world
have negative links. To solve this problem, signed GCNs aims to learn
graph representation with the additional signed link information.
Negative links usually contains semantic information that is different
from positive links, also the principles are inherently different from
positive links. The signed network will have a different representation
as $G = (V, E^+, E^-)$ where the signs of the edges are differentiated.
The aggregation for positive and negative links are different. Each
layer will have two representations, one for the balanced user where the
number of negative links is even and one for the unbalanced user where
the number of negative links is odd. The hidden states are
\begin{equation}\label{eq:hidden1}
h_i^{B(1)} = \sigma (W^{B(l)}[\sum \nolimits_{j \in N(i)^+} 
\frac{h_j^{0}}{|N_i^+|},h_i^{(0)}]),
\end{equation}
\begin{equation}\label{eq:hidden2}
h_i^{U(1)} = \sigma (W^{U(l)}[\sum \nolimits_{j \in N(i)^-} 
\frac{h_j^{0}}{|N_i^-|},h_i^{(0)}]),
\end{equation}
where $\sigma()$ is the non-linear activation function, $(W^{B(l)}$ and
$(W^{U(l)}$ are the linear transformation matrices for balanced and
unbalanced sets. 
\item Variational Graph Auto-Encoders (VGAE) \cite{kingma2013auto} \\
An autoencoder minimizes the reconstruction error of the input and the
output using an encoder and a decoder. The encoder maps input data to a
representation space. Then, it is further mapped to a reconstruction
space that preserves the neighborhood information.  VGAE uses GCN as
the encoder and an inner product decoder to embed graphs. 
\item GraphSAGE \cite{hamilton2017inductive} \\
GraphSAGE uses a sample and aggregate method to conduct inductive node
embedding with node features such as text attributes, node profiles,
etc. It trains a set of aggregation functions that integrate features of
local neighborhood and pass it to the target node $i$. Then, the hidden
state of node $i$ is updated by
\begin{equation}\label{eq:node}
h_i^{(k+1)} = ReLU(W^{(k)} h_i^{(k)}, \sum \nolimits_{n \in N(i)} 
(ReLU(Q^{(k)} h_n^{(k)}))),
\end{equation}
where $h_i^{(0)} = X_i$ is the initial node attributes and $\sum(.)$
denotes a certain aggregator function, {\em e.g.}, average, LSTM,
max-pooling, etc. 
\item Structural Deep Network Embedding (SDNE) \cite{wang2016structural} \\
SDNE learns a low-dimensional network-structure-preserving
representation by considering both the first-order and the second-order
proximities between vertexes using CNNs. To achieve this objective, it
adopts a semi-supervised model to minimize the following objective
function:
\begin{equation}\label{eq:obj}
{||(\hat{X}-X) \odot B||}_F^2 + \alpha \sum_{i,j = 1}^{n} s_{i,j} 
||y_i-y_j||_2^2 + v L_{reg},
\end{equation}
where $L_{reg}$ is an $L_2$-norm regularizing term to avoid overfitting,
$S$ is the adjacency matrix, and $B$ is the bias matrix. 
\end{itemize}

\subsection{Large Graph Embedding Methods}\label{LGEM}

To address the scalability issue, several embedding methods targeting at
large graphs have been proposed recently. They are examined in this subsection.

\begin{figure}[htbp]
\begin{center}
\includegraphics[width=7cm]{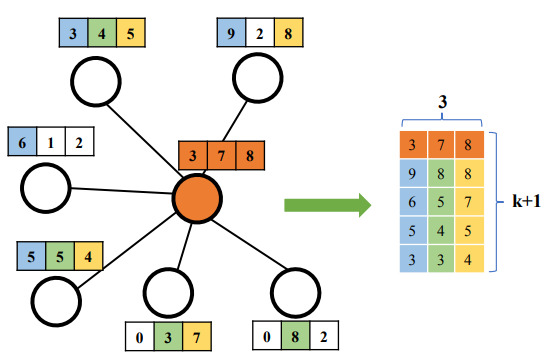}
\caption{Illustration of a learnable graph convolutional layer (LGCL) method
\cite{gao2018large}.}\label{fig:LGCL}
\end{center}
\end{figure}

\begin{figure}[htbp]
\begin{center}
\includegraphics[width=6.5cm]{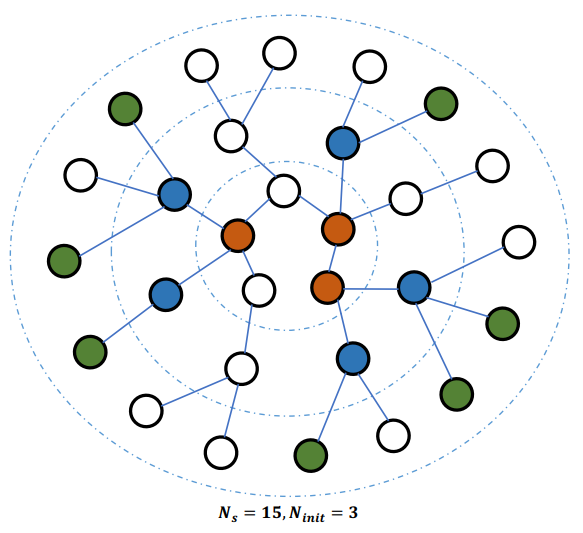}
\caption{Illustration of the sub-graph selection process \cite{gao2018large}.}
\label{fig:LGCL2}
\end{center}
\end{figure}

\begin{itemize}
\item Learnable graph convolutional layer (LGCL) \cite{gao2018large} \\
For each feature dimension, every node in the LGCL method selects a
fixed number of features from its neighboring nodes with value ranking.
Fig.  \ref{fig:LGCL} serves as an example. Each node in this figure has
a feature vector of dimension $n=3$. For the target node (in orange),
the first feature component of its six neighbors takes the values of 9,
6, 5, 3, 0, 0. If we set the window size to $k=4$, then the four largest
values (i.e., 9, 6, 5, 3) are selected.  The same process is repeated
for the two remaining features.  By including the feature vector of the
target node itself, we obtain a data matrix of dimension $(k + 1) \times
n$.  This results in a grid-like structure. Then, the traditional CNN
can be conveniently applied so as to generate the final feature vector.
To embed large-scale graphs, a sub-graph selection method is used to
reduce the memory and resource requirements. As shown in Fig.
\ref{fig:LGCL2}, it begins with $N_init = 3$ randomly sampled nodes (in
red) that are located in the center of the figure.  At the first
iteration, the breadth-first-search (BFS) is used to find all
first-order neighboring nodes of initial nodes.  Among them, $N_m = 5$
nodes (in blue) are randomly selected. At the next iteration, $N_m = 7$
nodes (in green) are randomly selected. After two iterations, 15 nodes
are selected as a sub-graph that serves as the input to LGCL. 

\item Graph partition neural networks (GPNN) \cite{motsinger2006gpnn} \\
GPNN extends graph neural networks (GNNs) to embed extremely large
graphs. It alternates between local (propagate information among nodes)
and global information propagation (messages among subgraphs). This
scheduling method can avoid deep computational graphs required by
sequential schedules. The graph partition is done using a multi-seed
flood fill algorithm, where nodes with large out-degrees are sampled
randomly as seeds. The subgraphs grow from seeds using flood fill,
which reaches out unassigned nodes that are direct neighbors of the 
current subgraph. 

\item LINE \cite{tang2015line} \\
LINE is used to embed graphs of an arbitrary type such as undirected,
directed and weighted graphs. It utilizes negative sampling to reduce
optimization complexity.  This is especially useful in embedding
networks containing millions of nodes and billions of edges. It is
trained to preserve the first- and second-order proximities, separately.
Then, the two embeddings are merged to generate a vector space to better
represent the input graph. One way to merge two embeddings is to
concatenate embedding vectors trained by two different objective
functions at each vertex. 

\end{itemize}

\subsection{Hypergraph Embedding}\label{HyperE}

Research on social network embedding grows quickly. A simple graph is
not powerful enough in representing the information of social networks.
The relation of vertices in social networks is far more complicated than
the vertex-to-vertex edge relationship.  Being different from
traditional graphs, edges in hypergraphs may have a degree larger than
two.  All related nodes are connected by a hyperedge to form a
supernode. Mathematically, an unweighted hypergraph is defined as
follows. A hypergraph, denoted by $G=(V,E)$, consists of a vertex set
$$
V = \{v_1,v_2,...,v_n\},
$$ 
and a hyperedge set, 
$$
E =\{e_1,e_2,...,e_m\}.
$$ 
A hyperedge, $e$, is said to be incident with a vertex $v$ if $v \in e$.
When $v \in e$, the incidence function $h(v,e) = 1$. Otherwise, $h(v, e)
= 0$. The degree of a vertex $v$ is defined as
$$
d(v) = \sum_{e \in E , v \in e} h(v, e).
$$ 
Similarly, the degree of a hyperedge $e$ is defined as 
$$
d(e) = \sum_{v \in V} h(v, e).
$$ 
A hypergraph can be represented by an incidence matrix $H$ of dimension
$|V| \times |E|$ with entries $h(v, e)$.  

\begin{figure}[htbp]
\includegraphics[width=8cm]{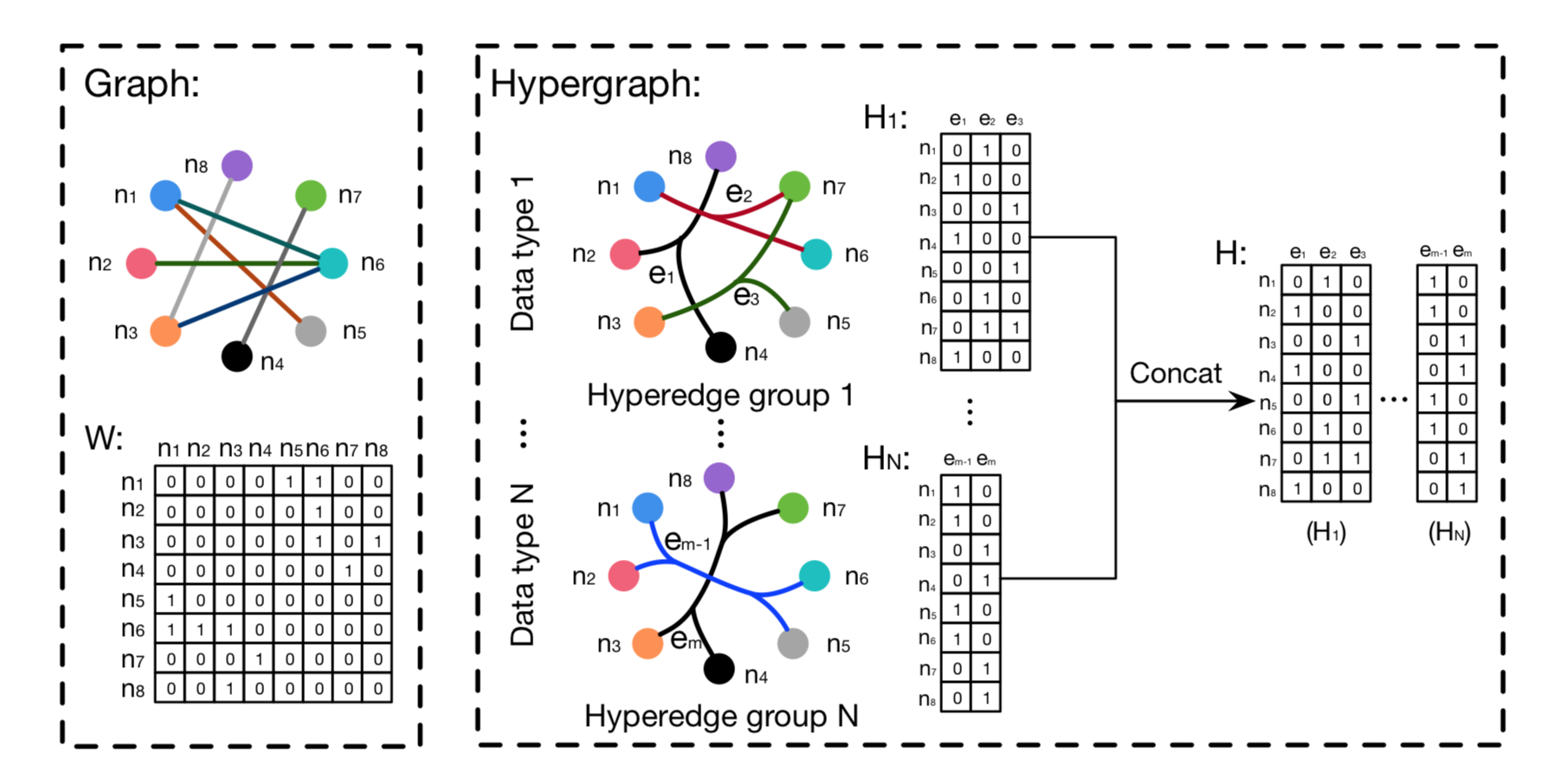}
\caption{Illustration of graph and hypergraph structures \cite{feng2018hypergraph}.}
\label{fig:comparison}
\end{figure}

Hyperedges possess the properties of edges and nodes at the same time.
As an edge, hyperedges connect multiple nodes that are closely related.
A hyperedge can also be seen as a supernode. For each pair of two
supernodes, their connection is established by shared incident vertices.
As a result, hypergraphs can better indicate the community structure in
the network data. These unique characteristics of hyperedges make
hypergraphs more challenging.  An illustration of graph and hypergraph
structures is given in Fig.  \ref{fig:comparison}.  It shows how to
express a hypergraph in table form.  The hyperedges, which are
indecomposable \cite{deep2017hypergraph}, can express the community
structure of networks.  Furthermore, properties of graphs and
hypergraphs are summarized and compared in Table \ref{table:comparison}.
Graphs and hypergraphs conversion techniques have been developed.
Examples include clique expansion and star expansion. Due to
indecomposibility of hyperedges, conversion from a hypergraph to a graph
will result in information loss. 

\begin{table}[htbp] 
\begin{center}
\begin{tabular}{|l|l|l|} \hline
                    & Graph       & Hypergraph \\ \hline
Representation      & A($|V|\times|V|$) & H($|V|\times|E|$) \\ \hline
Minimum Cut         & NP-Hard     & NP-Complete    \\ \hline
Spectral Clustering & Real-valued & Real-valued    \\ 
                    & optimization& optimization   \\ \hline
Spectral Embedding  & Matrix      & Project to     \\ 
                    & factorization & eigenspace   \\ \hline
\end{tabular}
\caption {Comparison of properties of graphs and hypergraphs.}\label{table:comparison}
\end{center}
\end{table}

Since hypergraphs provide a good tool for social network modeling, and
hypergraph embedding is a hot research topic nowadays.  On one hand,
hypergraph modeling has a lot of applications that are difficult to
achieve by graph modeling such as multi-modal data representation.  On
the other hand, hypergraphs can be viewed as a variant of simple graphs
and many graph embedding methods could be applied onto the hypergraphs
with minor modifications. There are embedding methods proposed for simple
graphs and they can be applied to hypergraphs as well as reviewed below. 
\begin{itemize}
\item Spectral Hypergraph Embedding \cite{zhou2007learning} \\
Hypergraph embedding can be treated as a k-way partitioning problem and
solved by optimizing a combinatorial function.  It can be further
converted to a real-valued minimization problem by normalizing the
hypergraph Laplacian. Its solution is any lower dimension embedding
space spanned by orthogonal eigenvectors of the hypergraph Laplacian,
$\Delta$, with the $k$ smallest eigenvalues.  
\item Hyper-Graph Neural Network (HGNN) \cite{feng2018hypergraph} \\
Being inspired by the spectral convolution on graphs in GCN
\cite{kipf2016semi}, HGNN applies the spectral convolution to
hypergraphs. By training the network through a semi-supervised node
classification task, one can obtain the node representation at the end
of convolutional layers. The architecture is depicted in Fig.
\ref{fig:HGNN}. The hypergraph convolution is derived from the
hypergraph Laplacian, $\Delta$, which is a positive semi-definite
matrix. Its eigenvectors provide certain basis functions while its
associated eigenvalues are the corresponding frequencies. The spectral
convolution in each layer is carried out via
\begin{equation}
f(X, W, \Theta) = \sigma(D_v^{\frac{-1}{2}} H W D_e^{-1} H^T 
D_v^{\frac{-1}{2}} X \Theta),
\end{equation}
where $X$ is the hidden embedding in each layer, $\Theta$ is the filter
response, and $D_v$ and $D_e$ are diagonal matrices with entries being
the degree of the vertices and the hyperedges, respectively. 
\begin{figure}[htbp]
\includegraphics[width=8cm]{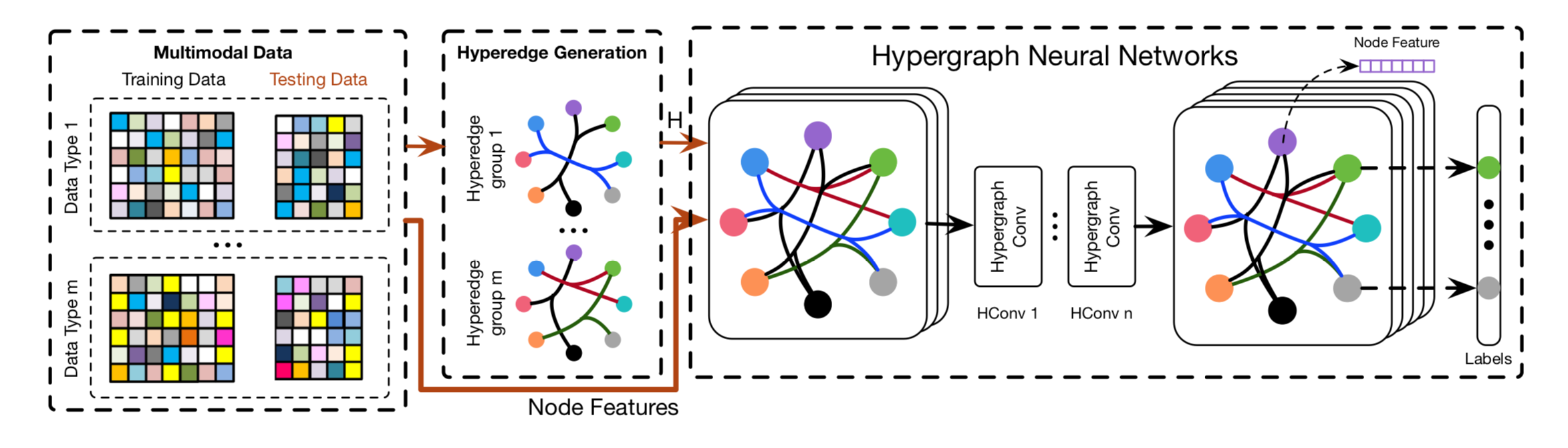}
\caption{The architecture of HGNN \cite{feng2018hypergraph}.}\label{fig:HGNN}
\end{figure}
\item Deep Hyper-Network Embedding (DHNE) \cite{deep2017hypergraph} \\
DHNE aims to preserve the structural information of hyperedges by a deep
neural auto-encoder. The auto-encoder first embed each vertex to a
vector in a lower dimensional latent space and then reconstruct it to
the original incidence vector afterwards. In the process of encoding and
decoding, the second order proximity is preserved to learn the global
structural information. The first order proximity is preserved in the
embedding space by defining an $N$-tuplewise similarity function. That
is, if $N$ nodes are in the same hyperedge, the similarity of these
nodes in the embedding space should be high.  Based on similarity, one
can predict whether $N$ nodes are connected by a single hyperedge.
However, the $N$-tuplewise similarity function should be non-linear;
otherwise, it will lead to contradicted predictions.  The local
information of a hypergraph can be preserved by shortening the distance
of connected vertices in the embedding space. 
\end{itemize}

\subsection{Attention Graph Embedding}\label{AGE}

An attention mechanism can be used in machine learning to allow the
learning process to focus on parts of a graph that are relevant to a
specific task.  One advantage of applying attention to graphs is to
avoid the noisy part of a graph so as to increase the signal-to-noise
(SNR) ratio \cite{lee2018attention} in information processing.
Attention-based node embedding aims to assign an attention weight,
$\alpha_i \in [0,1]$, to neighborhood nodes of a target node $t$, where
$\sum \nolimits_{i \in N(t)} \alpha_i = 1$ and $N(t)$ denotes the set of
neighboring nodes of $t$. 
\begin{itemize}
\item Graph Attention Networks (GAT) \cite{velivckovic2017graph} \\
GAT utilizes masked self-attentional layers to limit the shortcomings of
prior graph convolutional based methods. They aim to compute the
attention coefficients
\begin{equation}\label{eq:attention1}
\alpha _{ij} = \frac{\exp(LeakyReLU(\overrightarrow{a}^T[W
\overrightarrow{h_i}||W \overrightarrow{h_j}]))}{ \sum \nolimits_{k \in
N_i} \exp(LeakyReLU(\overrightarrow{a}^T[W \overrightarrow{h_i}||W
\overrightarrow{h_j}]},
\end{equation}
where $W$ is the weight matrix for the initial linear transformation,
then the transformed information on each neighbor's feature are
concatenated to obtain the new hidden state, which will be passed
through a $LeakyReLu$ activation. The above attention mechanism is a
single-layer feedforward neural network parameterized by the above
weight vector. 
\item AttentionWalks \cite{abu2017watch},\cite{abu2018watch} \\
Generally speaking, one can use the random walk to find the context of
the node.  For a graph, $G$, with corresponding transition matrix $T$
and window size $c$ the parameterized conditional expectation after a
$k$-step walk can be expressed as
\begin{equation}\label{eq:expectation}
E[D|q_1, q_2,...,q_c] = I_n \sum_{k=1}^{c} q_k T^k,
\end{equation}
where $I_n$ is the size-n identity matrix, $q_k$, $1 \leq i \leq c$, are
the trainable weights, $D$ is the walk distribution matrix whose entry
$D_{uv}$ encodes the number of times node $u$ is expected to visit node
$v$. The trainable weights are used to steer the walk towards a broader
neighborhood or restrict it within a smaller neighborhood.  Following
this idea, AttentionWalks adopts an attention mechanism to guide the
learning procedure. This mechanism suggests which part of the data to
focus on during the training process. The weight parameters are called
the attention parameters in this case. 
\item Attentive Graph-based Recursive Neural Network (AGRNN)
\cite{xu2017attentive}\\
AGRNN applies attention to a graph-based recursive neural network (GRNN)
\cite{xu2017collective} to make the model focus on vertices with more
relevant semantic information. It builds subgraphs to construct
recursive neural networks by sampling a number of $k$-step neighboring
vertices.  AGRNN finds a soft attention, $\alpha_r$, to control how
neighbor information should be passed to the target node. Mathematically, 
we have
\begin{equation}\label{eq:alpha}
\alpha_r = \mbox{Softmax} (x^T W^{(a)} h_r ),
\end{equation}
where $x_k$ is the input, $W^{(a)}$ is the weight to learn and
$h_r$ is the hidden state of the neighbors. 
The aggregated representation from all neighbors is used as the hidden
state of the target vertex
\begin{equation}\label{eq:vertex}
h_k = \sum\nolimits_{v_r \in N(v_k)} \alpha_r h_r,
\end{equation}
where $N(v_k)$ denotes the set of neighboring nodes of vertex $v_k$.
Although attention has been shown to be useful in improving some neural
network models, it does not always increase the accuracy of graph
embedding \cite{chen2018graph}. 
\end{itemize}

\subsection{Others}

\begin{itemize}
\item GraphGAN \cite{wang2018graphgan}  \\
GraphGAN employs both generative and discriminative models for graph
representation learning.  It adopts adversarial training and formulates
the embedding problem as a minimax game by borrowing the idea from the
Generative Adversarial Network (GAN) \cite{goodfellow2014generative}. To
fit the true connectivity distribution $p_true(v|v_c)$ of vertices
connected to target vertex $v_c$, GraphGAN models the connectivity
probability among vertices in a graph with a generator,
$G(v|v_c;\theta_G)$, to generate vertices that are most likely connected
to $v_c$.  A discriminator $D(v,v_c;\theta_D)$ outputs the edge
probability between $v$ and $v_c$ to differentiate the vertex pair
generated by the generator from the ground truth. The final vertex
representation is determined by alternately maximizing and minimizing
the value function $V(G,D)$ as
\begin{equation}\label{eq:minimax}
\begin{split}
\min_{\theta_G}\max_{\theta_D} V(G,D)\\
V(G,D) = \sum_{c=1}^V (\mathbb{E}_{v \sim p_{true} (\cdot|v_c)[\log D(v,v_c;\theta_D)])} \\
+\mathbb{E}_{v \sim G (\cdot|v_c;\theta_G)}[\log( 1-D(v,v_c;\theta_D)]).
\end{split}
\end{equation} 
\item GenVector \cite{yang2015multi} \\
GenVector leverages large-scale unlabeled data to learn large social
knowledge graphs. It is cast as a weakly supervised problem and solved by
unsupervised techniques with a multi-modal Bayesian embedding model.
GenVector can serve as a generative model in applications. For example,
it uses latent discrete topic variables to generate continuous word
embeddings and graph-based user embeddings and integrates the advantages
of topic models and word embeddings. 
\end{itemize}

\section{Evaluation}\label{sec:evaluation}

We study the evaluation of various graph representation methods in this
section. Evaluation tasks and datasets will be discussed in Secs.
\ref{subsec:tasks} and \ref{subsec:datasets}, respectively. Then,
evaluation results will be presented and analyzed in Sec.
\ref{subsec:results}. 

\subsection{Evaluation Tasks}\label{subsec:tasks}

The two most common evaluation tasks are vertex classification and link
prediction. We use vertex classification to compare different graph
embedding methods and draw insights from the obtained results. 
\begin{itemize}
\item Vertex Classification \\
Vertex classification aims to assign a class label to each node in a
graph based on the information learned from other labeled nodes.
Intuitively, similar nodes should have the same label.  For example,
closely-related publication may be labeled as the same topic in the
citation graph while individuals of the same gender, similar age and
common interests may have the same preference in social networks.  Graph
embedding methods embed each node into a low-dimensional vector. Given
an embedded vector, a trained classifier can predict the label of a
vertex of interest, where the classifier can be SVM (Support Vector
Machine) \cite{golub1971singular}, logistic regression
\cite{wang2017community}, kNN (k nearest neighbors)
\cite{le2014probabilistic}, etc.  The vertex label can be obtained in an
unsupervised or semi-supervised way. Node clustering is an unsupervised
method that groups similar nodes together.  It is useful when labels are
unavailable.  The semi-supervised method can be used when part of the
data are labeled. The F1 score is used for evaluation in binary-class
classification while the Micro-F1 score is used in multi-class
classification. Since accurate vertex representations contribute to high
classification accuracy, vertex classification can be used to measure
the performance of different graph embedding methods.  
\item Link Prediction \cite{gao2011temporal} \\
Link prediction aims to infer the existence of relationship or
interaction among pairs of vertices in a graph.  The learned
representation should help infer the graph structure, especially when
some links are missing. For example, links might be missing between two
users and link prediction can be used to recommend friends in social
networks.  The learned representation should preserve the network
proximity and the structural similarity among vertices.  The information
encoded in the vector representation for each vertex can be used to
predict missing links in incomplete networks. The link prediction
performance can be measured by the area under curve (AUC) or the
receiver operating characteristic (ROC) curve.  A better representation
should be able to capture the connections among vertices better. 
\end{itemize}
We describe the benchmark graph datasets and conduct experiments in
vertex classification on both small and large datasets in the following
subsections. 

\subsection{Evaluation Datasets}\label{subsec:datasets} 

Citation datasets such as Citeseer \cite{giles1998citeseer}, Cora
\cite{cabanes2013cora} and PubMed \cite{canese2013pubmed} are examples
of small datasets. They can be represented as directed graphs in 
which edges indicates author-to-author or paper-to-paper citation
relationship and text attributes of paper content at nodes.  

First, we describe several representative citation datasets below. 
\begin{itemize}
\item Citeseer \cite{giles1998citeseer} \\ 
It is a citation index dataset containing academic papers of six
categories. It has 3,312 documents and 4,723 links.  Each document is
represented by a 0/1-valued word vector indicating the absence/presence
of the corresponding word from a dictionary of 3,703 words. Thus, the
text attributes of a document is a binary-valued vector of 3,703
dimensions. 
\item Cora \cite{cabanes2013cora} \\
It consists of 2,708 scientific publications of seven classes.  The
graph has 5,429 links that indicate citation relations between
documents. Each document has text attributes that are expressed by a
binary-valued vector of 1,433 dimensions. 
\item WebKB \cite{trochim2001research} \\
It contains seven classes of web pages collected from computer science
departments, including student, faculty, course, project, department,
staff, etc. It has 877 web pages and 1,608 hyper-links between web
pages. Each page is represented by a binary vector of 1,703 dimensions. 
\item KARATE \cite{zachary1977information} \\
Zachary's karate network is a well-known social network of a university
karate club. It has been widely studied in social network analysis. The
network has 34 nodes, 78 edges and 2 communities. 
\item Wikipedia \cite{cucerzan2007large} \\
The Wikipedia is an online encyclopedia created and edited by volunteers
around the world.  The dataset is a word co-occurrence network constructed from
the entire set of English Wikipedia pages. 
This data contains 2405 nodes, 17981 edges and 19 labels. 
\end{itemize}

Next, we present several commonly used large graph datasets below. 
\begin{itemize}
\item Blogcatalog \cite{tang2012unsupervised} \\
It is a network of social relationships of bloggers listed in the
BlogCatalog website. The labels indicate blogger's interests inferred
from the meta-data provided by bloggers. The network has 10,312 nodes,
333,983 edges and 39 labels. 
\item Youtube \cite{wattenhofer2012youtube} \\
It is a social network of Youtube users. This graph contains 1,157,827
nodes, 4,945,382 edges and 47 labels. The labels represent groups of
users who enjoy common video genres. 
\item Facebook \cite{lewis2008tastes} \\ 
It is a set of postings collected from the Facebook website.  This data
contains 4039 nodes, 88243 edges and no labels. It is used for link
prediction. 
\item Flickr \cite{sigurbjornsson2008flickr} \\
It is an online photo management and sharing dataset.  It contains 80513
nodes, 5899882 edges and 195 labels. 
\end{itemize}

Finally, parameters of the above-mentioned datasets are summarized in 
Table \ref{table:datasum}. 
\begin{table}[htbp] 
\begin{center}
\begin{tabular}{|l|l|l|l|}\hline
            & Node    & Edge    & Labels \\ \hline
Citeseer    & 3312    & 4723    & 6      \\ \hline
Cora        & 2708    & 5429    & 7      \\ \hline
WebKB       & 877     & 1608    & 7      \\ \hline
KARATE      & 34      & 78      & N/A    \\ \hline
Wiki        & 2405    & 17981   & 19     \\ \hline
Blogcatalog & 10312   & 333983  & 39     \\ \hline
Youtube     & 1157827 & 4945382 & 47     \\ \hline
Facebook    & 4039    & 88243   & N/A    \\ \hline
Flickr      & 80513   & 5899882 & 195    \\ \hline
\end{tabular}
\caption {Summary of representative graph datasets.}\label{table:datasum}
\end{center}
\end{table}

\subsection{Evaluation Results and Analysis}\label{subsec:results} 

Since evaluations were often performed independently on different
datasets under different settings in the past, it is difficult to draw a
concrete conclusion on the performance of various graph embedding
methods. Here, we compare the performance of graph embedding methods
using a couple of metrics under the common setting and analyze obtained
results. In addition, we provide an open-source Python library, called
the Graph Representation Learning Library
(\href{https://github.com/jessicachen626/GRLL}{GRLL}), to readers in the
Github. It offers a unified interface for all graph embedding methods
that were experimented in this work.  To the best of our knowledge, this
library covers the largest number of graph embedding techniques up to
now. 

\subsubsection{Vertex Classification}

We compare vertex classification accuracy of seven graph embedding
methods on Cora and Wiki. We used the default hyper-parameter setting
provided by each graph embedding method. For the classifier, we adopted
linear regression for all methods. We split samples equally into the
training and the testing sets (i.e. 50\% and 50\%).  The vertex
classification results are shown in Table \ref{tab:cora}.  DeepWalk and
node2vec offer the highest accuracy for Cora and Wiki, respectively.
The random-walk-based methods (e.g., DeepWalk, node2vec and GraRep) are
the top three performers for both Cora and Wiki. DeepWalk and node2vec
are preferred among them since GraRep usually demands much more memory
and it is difficult to apply GraRep to larger graphs. 

\begin{table}[htbp]
\begin{center}
\begin{tabular}{|c|c|c|}\hline
         & Cora   & Wiki \\ \hline
DeepWalk & \textbf{0.829}  & 0.670\\ \hline
node2vec & 0.803  & \textbf{0.680}\\ \hline
GraRep   & 0.788  & 0.650\\ \hline
HOPE     & 0.646  & 0.608\\ \hline
SDNE     & 0.573  & 0.510\\ \hline
LINE     & 0.432  & 0.520\\ \hline
GF       & 0.499  & 0.465\\ \hline
\end{tabular}
\caption{Performance comparison of seven graph embedding methods in 
vertex classification on Cora and Wiki.}\label{tab:cora}
\end{center}
\end{table}

\begin{table*}[htb]
\begin{center}
\begin{tabular}{|c|c|c|c|c|c|c|c|}
\hline
 & & DeepWalk & node2vec & LINE & GraRep & GF & HOPE \\ \hline
YouTube & Macro-F1 & 0.206 & \textbf{0.221} & 0.170 & N/A & N/A & N/A \\
 & Micro-F1 & 0.293 & \textbf{0.301} & 0.266 & N/A & N/A & N/A \\ \hline
Flickr & Macro-F1 & \textbf{0.212} & 0.203 & 0.162 & N/A & N/A & N/A\\ 
 & Micro-F1 & \textbf{0.313} & 0.311 & 0.289 & N/A & N/A & N/A \\ \hline
BlogCatalog & Macro-F1 & 0.247 & \textbf{0.250} & 0.194 & 0.230 & 0.072 & 0.143 \\
 & Micro-F1 & 0.393 & \textbf{0.400} & 0.356 & 0.393 & 0.236 & 0.308  \\ \hline
\end{tabular}
\caption{Comparison of clustering quality of six graph embedding methods in terms of 
Macro- and Micro-F1 scores against three large graph datasets.}\label{tab:Clust}
\end{center}
\end{table*}

\subsubsection{Clustering Quality}

We compare various graph embedding methods by examining their clustering
quality in terms of the Macro- and Micro-F1 scores. The K-means++
algorithm is adopted for the clustering task. Since the results of
K-means++ clustering are dependent upon seed initialization, we perform
10 consecutive runs and report the best result. We tested them on three
large graph datasets (i.e., YouTube, Flickr and BlogCatalog).  The
experimental results are shown in Table \ref{tab:Clust}.  YouTube and
Flickr contain more than millions of nodes and edges and we can only run
DeepWalk, node2vec and LINE on them with the 24G RAM limit as reported
in the table.  We see that DeepWalk and node2vec provide the best
results. They are both random-walk based methods with different sampling
scheme. Also, they demand less memory as compared with others.  In
general, random walk with the skip-gram model is a good baseline for
unsupervised graph embedding. GraRep offers comparable graph embedding
quality for BlogCatalog. However, its memory requirement is huge so that
it is not suitable for large graphs. 

\subsubsection{Time Complexity}
    
\begin{table}[htb]
\begin{center}
\begin{tabular}{|l|l|l|l|}\hline
           & YouTube    & Flickr    & Wiki      \\ \hline
DeepWalk    & 37366       & 3636.14    & 37.23 \\ \hline
node2vec     & 41626.94     & 40779.22    & 27.53  \\ \hline
LINE       & 185153.29       & 31707.87      & 79.42     \\ \hline
\end{tabular}
\caption {Comparison of time used in training (seconds).}\label{tab:timecomplexity}
\end{center}
\end{table}

Time complexity is an important factor to consider, which is especially
true for large graphs.  The time complexity of three embedding methods
against three datasets is compared in Table \ref{tab:timecomplexity}.
We see that the training time of DeepWalk is significantly lower than
node2vec and LINE for larger graph datasets such as YouTube and Flickr.
DeepWalk is an efficient graph embedding method with high accuracy by
considering embedding quality as well as training complexity. 

\subsubsection{Influence of Embedding Dimensions}

\begin{figure}[htbp] 
\includegraphics[width=8cm]{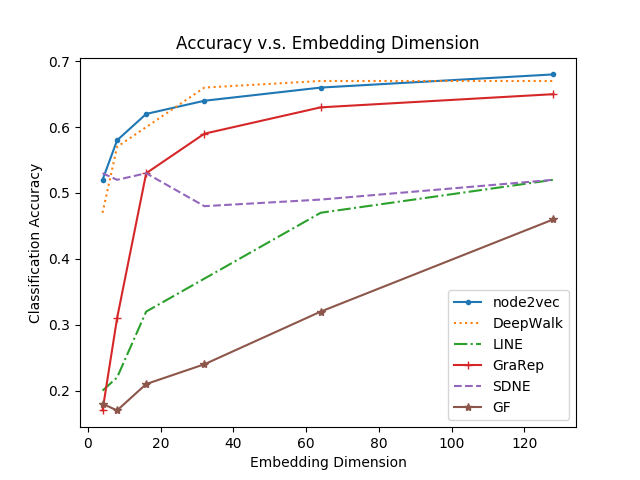}
\caption{The node classification accuracy as a function of the embedding dimension
for the Wiki dataset.}\label{fig:embedding_dim} 
\end{figure}

As the embedding dimension decreases, less information of the input
graph is preserved so that the performance drops. However, some drops
faster than others. We show the node classification accuracy as a
function of the embedding dimension for the Wiki dataset in Fig.
\ref{fig:embedding_dim}. We compare six graph embedding methods
(node2vec, DeepWalk, LINE, GraRep, SDNE and GF) and their embedding
dimensions vary from 4, 8, 16, 32, 64 to 128. We see that the performance of the random-walk based embedding methods (node2vec and Deep-Walk) degrades slowly. 
Only about 20\% drop in performance when the embedding dimension
size goes from 128 to 4. In contrast, The performance of the structural
preserving methods (LINE and GreRep) and the graph factorization method
(GF) drops significantly (as much as 45\%) when the embedding size goes
from 128 to 4. One explanation is that the structural preserving methods
optimize the representation vectors in the embedding space so that a
small information loss will result in substantial difference.
Random-walk based methods obtain embedding vectors by selecting paths
from the input graph randomly. Yet, the relationship between nodes is
still preserved when the embedding dimension is small. SDNE adopts the
auto-encoder architecture to preserve the information of the input graph
so that its performance remains about the same regardless of the
embedding dimension. 
    
\begin{figure}[htbp]
\includegraphics[width=8cm]{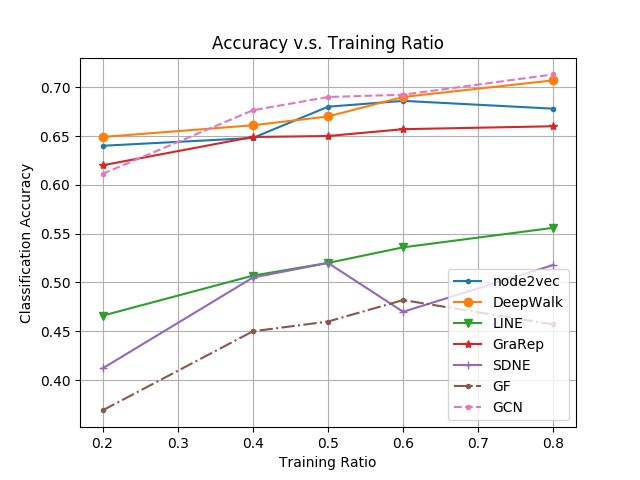}
\caption{The node classification accuracy as a function of the training
sample ratio for the Cora dataset.}\label{fig:ratio}
\end{figure}

\subsubsection{Influence of Training Sample Ratio}

By the training sample ratio, we mean the percentages of total graph
samples that are used for the training purpose.  When the ratio is high,
the classifier could be overfit. On the other hand, if the ratio is too
low, the offered information may not be sufficient for the training
purpose. Such analysis is classifier dependent, and we adopt a simple
linear regression classifier from the python sklearn toolkit in the
experiment. The node classification accuracy as a function of the
training sample ratio for the Cora dataset is shown in Fig.
\ref{fig:ratio}.  Most methods have consistent performance for the
training data ratio between 0.2 and 0.8 except for the machine learning
based methods (SDNE and GCN). Their accuracy drops when the training
data ratio is low. They need a larger number of training data. 
    
\section{Emerging Applications}\label{sec:applications}

Graphs offer a powerful modeling tool and find a wide range of
applications. Since many real-world data have certain relationship
between entities, they can be conveniently modeled by graphs.
Multi-modal data can also be embedded into the same space through graph
representation learning and, as a result, the information from different
domains can be represented and analyzed in one common setting. 

In this section, we examine three emerging areas that benefit from graph
embedding techniques. 
\begin{itemize}
\item Community Detection \\
Graph embedding can be used to predict the label of a node given a
fraction of labeled node \cite{chen2009novel}, \cite{ding2001min},
\cite{yuruk2009ahscan}, \cite{zhang2015finding}. Thus, it has been
widely used for community detection \cite{dourisboure2007extraction},
\cite{newman2004detecting}. In social networks, node labels might be
gender, demography or religion. In language networks, documents might be
labeled with categories or keywords. Missing labels can be inferred from
labeled nodes and links in the network.  Graph embedding can be used to
extract node features automatically based on the network structure and
predict the community that a node belongs to. Both vertex classification
\cite{girvan2002community} and link prediction \cite{de2017community}
\cite{zheleva2008using} can facilitate community detection 
\cite{fortunato2010community}, \cite{yang2009combining}. 
\item Recommendation System \\
Recommendation is an important function in social networks and
advertising platforms \cite{hamilton2017representation},
\cite{ying2018graph}, \cite{zhang2016collaborative}. Besides the
structure, content and label data \cite{kermarrec2011distributed}, some
networks contain spatial and temporal information. For example, Yelp may
recommend restaurants based on user's location and preference.
Spatial-temporal embedding \cite{zhang2017regions} is an emerging topic
in mobile applications. 
\item Graph Compression and Coarsening \\
By graph compression (graph simplification), we refer to a process of
converting one graph to another, where the latter has a smaller number
of edges. It aims to store a graph more efficiently and run graph
analysis algorithms faster. For example, a graph is partitioned into
bipartite cliques and replaced by trees to reduce the edge number in
\cite{feder1995clique}.  Along this line, one can also aggregate nodes
or edges for faster processing with graph coarsening
\cite{liang2018mile}, where a graph is converted into smaller ones
repeatedly using a hybrid matching technique to maintain its backbone
structure.  The Structural Equivalence Matching (SEM) method
\cite{green2009reliability} and the Normalized Heavy Edge Matching
method (NHEM) \cite{karypis1998multilevelk} are two examples. 
\end{itemize}

\section{Future Research Directions}\label{sec:future}

Several future research opportunities in graph embedding are discussed below. 
\begin{itemize}
\item Deep Graph Embedding \\
GCN \cite{kipf2016semi} has drawn a lot of attention due to its superior
performance. However, the number of graph convolutional layers in
typically not greater than two. When there are more graph convolutional
layers in cascade, it is surprising that the performance drops
significantly.  It was argued in \cite{li2018deeper} that each GCN layer
corresponds to graph Laplacian smoothing since node features are
propagated in the spectral domain.  When a GCN is deeper, the graph
Laplacian is over-smoothed and the corresponding node features become
obscure.  Yet, each layer of GCN only learns one-hop information, and
two GCN layers learn the first and second-order proximity in the graph.
It is difficult for a shallow structure to learn the global information.
The receptive field of each filter in GCN is global since graph
convolution is conducted in the spectral domain.  One solution to fix
this problem is to conduct the convolution in the spatial domain.  For
example, one can convert graph data into grid-structure data as proposed
in \cite{gao2018large}. Then, the graph representation can be learned
using multiple CNN layers.  Another way to address the problem is to
down-sample graphs and merge similar nodes together. Then, we can build
a hierachical network structure, which allows to learn both local and
global graph data in a hierarchical manner. Such a graph coarsening idea
was adopted by \cite{gao2019learning} \cite{hu2019semi},
\cite{ying2018hierarchical} to build deep GCNs. 
\item Semi-supervised Graph Embedding \\
Classical graph embedding methods such as PCA, DeepWalk and matrix
factorization are unsupervised learning methods. They use topological
structures and node attributes to generate graph representations. No
labels are required in the training process.  However, labels are useful
since they provide more information about the graph. In most real world
applications, labels are available in a subset of nodes, leading to a
semi-supervised learning problem.  The feasibility for semi-supervised
learning on graphs was discussed in \cite{yang2016revisiting},
\cite{zhou2004learning}. A large number of graph embedding problems
belong to the semi-supervised learning paradigm, and it deserves our
attention.
\item  Dynamic Graph Embedding \\
Social graphs such as the twitter are constantly changing. Another
example is graphs of mobile users whose location information is changing
along with time. To learn the representation of dynamic graphs is an
important research topic and it finds applications in real-time and
interactive processes such as the optimal travel path planning in a city
at traffic hours. 
\item Scalability of Graph Embedding \\
We expect to see graphs of a larger scale and higher diversity because
of the rapid growth of social networks, which contain millions and
billions of nodes and edges. It is still an open problem how to embed
very large graph data efficiently and accurately.  
\item Interpretability of Graph Embedding \\
Most state-of-the-art graph embedding methods are built upon CNNs, which
are trained with backpropagation (BP) to determine their model
parameters.  However, the training complexity is very high.  Besides,
CNNs are mathematically intractable. Research was done to lower the
training complexity such as quickprop \cite{fahlman1988empirical}.
Furthermore, there is new work \cite{kuo2018interpretable} that attempts
to explain CNNs using an interpretable and feedforward (FF) design
without any BP.  The work in \cite{kuo2018interpretable} adopts a
data-centric approach to network parameters of the current layer based
on data statistics from the output of the previous layer in a one-pass
manner. 
\end{itemize}

\section{Conclusion}\label{sec:conclusion}

A comprehensive survey of the literature on graph representation
learning techniques was conducted in this paper.  We examined various
graph embedding techniques that convert the input graph data into a
low-dimensional vector representation while preserving intrinsic graph
properties. Besides classical graph embedding methods, we covered
several new topics such as neural-network-based embedding methods,
hypergraph embedding and attention graph embedding methods.
Furthermore, we conducted an extensive performance evaluation of several
stat-of-the-art methods against small and large datasets.  For
experiments conducted in our evaluation, an open source Python library,
called the Graph Representation Learning Library (GRLL), was provided to
readers. Finally, we presented some emerging applications and future
research directions. 

\bibliography{graph_embedding_survey}

\end{document}